\documentclass[runningheads]{llncs}
\usepackage{graphicx}
\usepackage[hidelinks]{hyperref}
\usepackage{booktabs}
\usepackage{amsmath}
\usepackage{amssymb}
\usepackage{multirow}
\usepackage{array}
\usepackage[numbers]{natbib}
\usepackage{etoolbox}

\newbool{anon}
\setbool{anon}{false}

\begin{document}
\title{Pose2Gait: Extracting Gait Features from \\Monocular Video of Individuals with Dementia}
\ifbool{anon}
{ 
    \author{Anonymous}
    \institute{
        Anonymous Organization \\
        \email{***@***.***}
    }
}
{ 
    \author{
        Caroline Malin-Mayor\inst{1,2} \and
        Vida Adeli\inst{1,2} \and
        Andrea Sabo\inst{2} \and
        Sergey Noritsyn\inst{1} \and
        Carolina Gorodetsky \inst{3} \and
        Alfonso Fasano \inst{1,2,4,5,6} \and
        Andrea Iaboni \inst{1,2,7} \and
        Babak Taati \inst{1,2}
    }
    \institute{
        University of Toronto
        \and 
        KITE Research Institute, Toronto Rehabilitation Institute, University Health Network
        \and
        The Hospital for Sick Children
        \and
        Morton and Gloria Shulman Movement Disorders Clinic, Toronto Western Hospital, University Health Network
        \and
        Krembil Brain Institute, University Health Network
        \and
        Center for Advancing Neurotechnological Innovation to Application
        \and
        Centre for Mental Health, University Health Network
        \\ \vspace{6pt}
        Toronto, Ontario, Canada
    }
}
        

\titlerunning{Pose2Gait: Extracting Gait Features from Monocular Video}

\ifbool{anon}
    {\authorrunning{Anonymous et al.}} 
    {\authorrunning{C. Malin-Mayor et al.}}

\maketitle              
\begin{abstract}
  Video-based ambient monitoring of gait for older adults with dementia has the potential to detect negative changes in health and allow clinicians and caregivers to intervene early to prevent falls or hospitalizations.
  Computer vision-based pose tracking models can process video data automatically and extract joint locations; however, publicly available models are not optimized for gait analysis on older adults or clinical populations. 
  In this work we train a deep neural network to map from a two dimensional pose sequence, extracted from a video of an individual walking down a hallway toward a wall-mounted camera, to a set of three-dimensional spatiotemporal gait features averaged over the walking sequence. 
  The data of individuals with dementia used in this work was captured at two sites using a wall-mounted system to collect the video and depth information used to train and evaluate our model. 
   Our Pose2Gait model\footnote{Code available at \href{https://github.com/TaatiTeam/pose2gait_public}{https://github.com/TaatiTeam/pose2gait\_public}} is able to extract velocity and step length values from the video that are correlated with the features from the depth camera, with Spearman's correlation coefficients of .83 and .60 respectively, showing that three dimensional spatiotemporal features can be predicted from monocular video. 
   Future work remains to improve the accuracy of other features, such as step time and step width, and test the utility of the predicted values for detecting meaningful changes in gait during longitudinal ambient monitoring.
\end{abstract}

\section{Introduction}
Ambient monitoring of gait has the potential to be a powerful tool for improving quality of life for individuals with dementia. Not only do individuals with dementia fall at a rate at least two times higher than cognitively healthy individuals \cite{harlein_2009}, but changes in gait can reflect injuries or illnesses that individuals with dementia have trouble communicating to caregivers. Daily ambient monitoring of gait features such as velocity and step length could quickly and automatically detect changes that indicate serious underlying issues, allowing early intervention and treatment. While historically gait analysis has required costly motion capture systems, recent advances in computer vision research have made it possible to reconstruct two-dimensional or three-dimensional poses from a simple monocular video. These computer vision-based pose tracking models open up the possibility for new clinical applications to detect fall risk or gait anomalies based on ambient monitoring of gait with an inexpensive wall-mounted video camera. 

Yet generic computer vision research does not optimize pose trackers for gait analysis, which has a small range of common poses but requires precise lower body joint placement. Additionally, there is a gap between having joint locations from a pose estimator and knowing clinically relevant gait features such as step time, step length, and velocity. With knowledge of heel down and toe off times, one can heuristically extract gait features from pose sequences, but automatic heel strike detection is highly dependent on the quality of the pose sequences. Therefore, small errors in pose estimation can cause large errors in step time and extracted gait features. Training a neural network to predict gait features from pose sequences provides a more flexible approach. This work shows how a neural network can be trained to predict three-dimensional gait features from two-dimensional pose sequences extracted from monocular video of individuals with dementia with open-source pose estimators.

\section{Related Work}

Monocular human pose estimation and tracking in two and three dimensions is an active area of computer vision research, with increasingly many methods and benchmark datasets. Surveys  \cite{liu_recent_2022, zhang_single_2021} provide an overview of the field. With the increasing capability of these methods to extract pose information from a single image or video, there is also increasing research on applying them to clinical use cases, both extracting gait features and predicting downstream clinical values. 

Previous work has validated the use of computer vision-based pose tracking to extract temporal and spatiotemporal gait features from monocular video. 
Stenum et al. \cite{stenum_two-dimensional_2021} validate gait features extracted from saggital plane videos using OpenPose against motion capture values and find high correlation for step time, stance time, swing time, double support time, and step length. They also indicate that positioning of the person with respect to the camera affected the accuracy of results, with highest accuracy when the person was directly in front of the camera and lower accuracy when they were approaching or past the camera. 
Sabo et al. \cite{sabo_concurrent_2022} validate gait features extracted from a variety of pose tracking models against a Zeno\textsuperscript{TM} Instrumented Walkway during clinical analysis of individuals with Parkinson's disease. The videos in this dataset had a frontal view of the body, with the participant walking toward or away from the camera. The analysis showed a moderate to strong positive correlation for number of steps, cadence, and step width between the Zeno\textsuperscript{TM} Walkway and the two-dimensional pose tracking models with automated heel strike detection.
Correlations were low for step time, and features such as step length and velocity could not be computed using this extraction method from two-dimensional poses. 
The results from these validation studies illustrate that while pose tracking has promise for gait analysis, further work is required to extract features with high accuracy in challenging clinical conditions.

Lonini et al. \cite{lonini_2022} and Cotton et al. \cite{cotton_transforming_2022} address these challenges by training an additional model to extract gait features from pose sequences. Lonini et al. \cite{lonini_2022} use DeepLabCut to train a custom pose tracker focusing on lower body keypoints in below-waist videos of stroke survivors. They then trained a convolutional neural network to predict the mean value over the walk for temporal gait features including cadence and per-leg swing and stance time, and compare the accuracy to a GAITRite\textsuperscript{\textregistered} walkway, achieving correlations of greater than .9 for all features except swing time.  Cotton et al. \cite{cotton_transforming_2022} also use a neural network to predict gait features from pose sequences on a diverse clinical dataset with a median age of 11 years and cerebral palsey and spina bifida as the most common conditions. The authors first use open-source pose tracking models to extract three-dimensional pose sequences from video. They then train a transformer neural network to predict hip and knee flexion angle, forward position of foot relative to pelvis, forward velocity of feet and pelvis, and step timing information.
They compare this neural network to a motion capture and force plate system and achieve high correlations for cadence (0.98), step time (0.95), step length (0.70), velocity (0.88), single support time (0.71), and double stance time (0.91). 

In addition to prior work extracting gait features from monocular video, there is also significant literature on predicting clinical outcomes from gait features. Gait features computed from monocular video using OpenPose are used to predict UPDRS scores and detect anti-psychotic induced Parkinsonism in individuals with dementia \cite{sabo_assessment_2020}. Gait features can also be used to predict short-term fall risk in high-risk individuals from 2D video \cite{ng_measuring_2020} and depth camera data \cite{mehdizadeh_predicting_2021}.
Aich et al. \cite{aich_validation_2018} use gait features extracted from accelerometer data to predict freezing of gait in individuals with Parkinson's disease. These clinical tools were validated using gait features from a variety of sources, such as motion capture systems, wearable accelerometers, and depth cameras, each of which is more expensive and less accessible than a simple video camera. Accurately predicting gait features from monocular video will increase access to these clinical tools beyond labs with specialized equipment.

\section{Methods}
\label{sec:methods}

\begin{figure}
    \centering
    \includegraphics[width=.7\columnwidth]{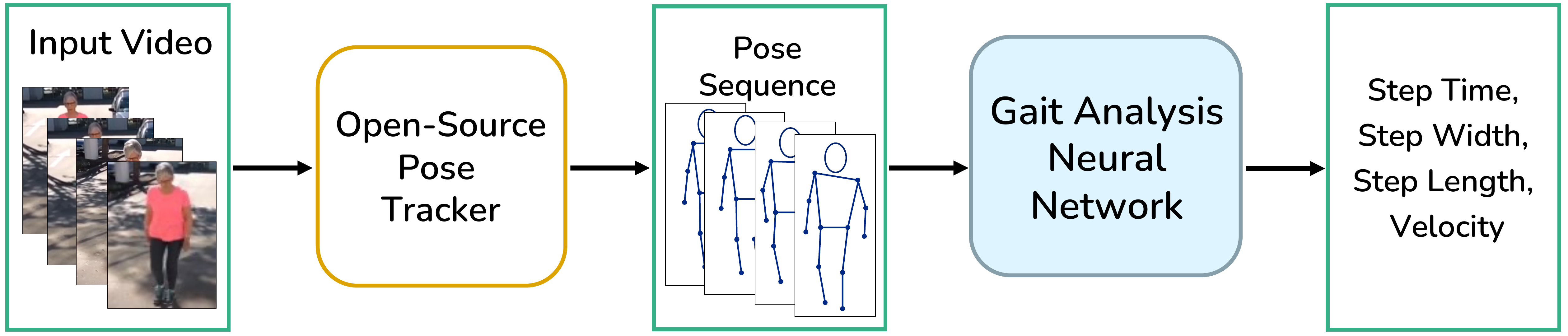}
    \caption{Pipeline for predicting gait features from monocular video. The input is a frontal video of an individual walking toward the camera. An open-source two-dimensional pose estimator is used to predict joint locations for each frame in the video. After interpolation and smoothing, the two-dimensional pose sequences are fed into a gait feature predictor neural network, which outputs a set of four gait features for the walk, representing the mean value over all steps in the walk.}
    \label{fig:overview}
\end{figure}

\begin{figure}
    \centering
    \includegraphics[width=.6\columnwidth]{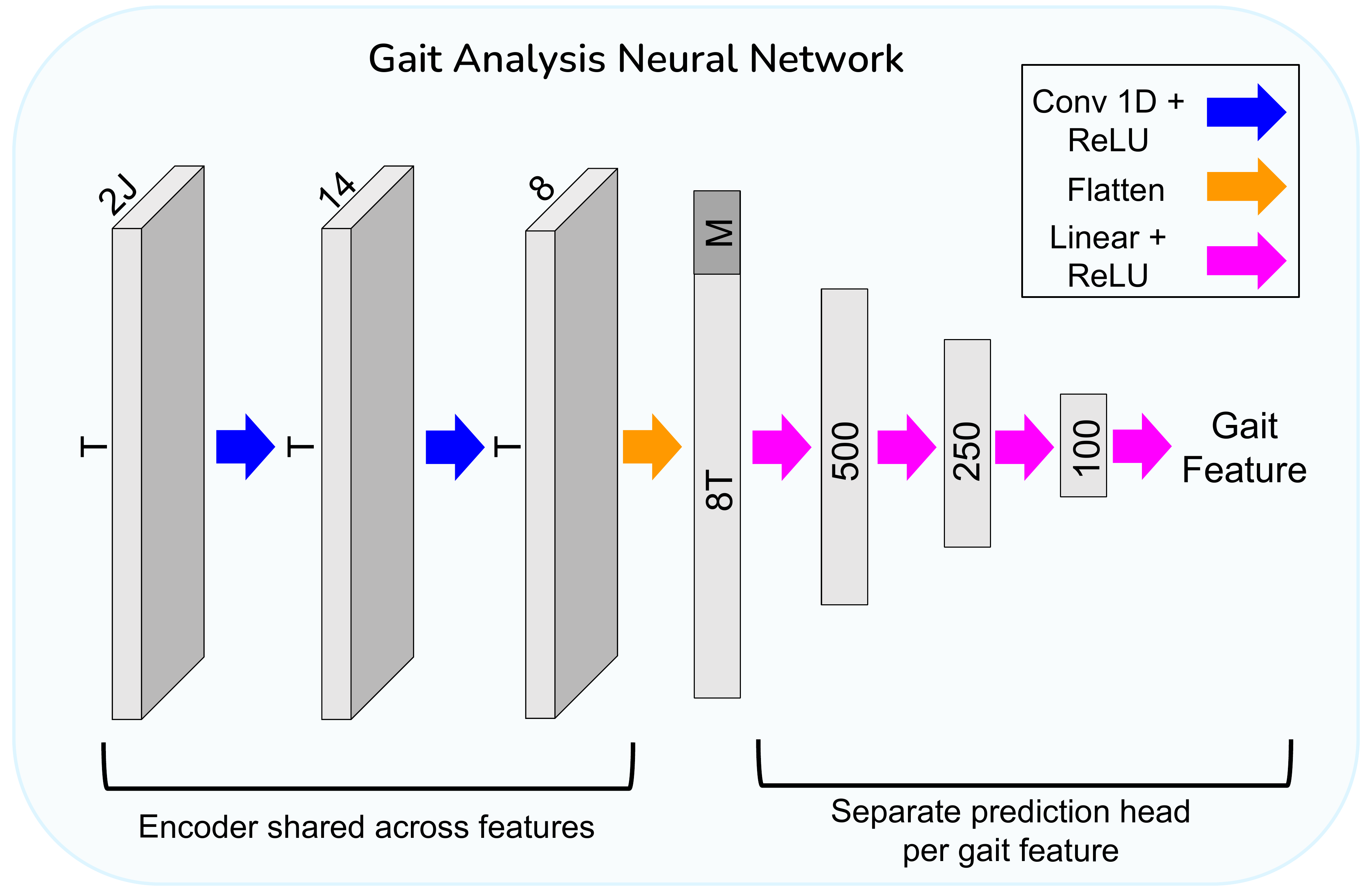}
    \caption{Architecture of the gait feature predictor neural network. It consists of two sections, an encoder which is shared by all gait features, and a separate prediction head for each gait feature. The encoder uses multiple layers of 1D convolutions to incorporate information over the time dimension, while the prediction head uses simple linear layers. Here, T represents the number of input frames in the pose sequence, while J is the number of joints. Additional metadata of length M is also added before the prediction head. This includes one-hot encodings representing the two-dimensional pose tracker used to generate the sequence and the dataset that the sequence is from. In our experiments, T=120 (4 seconds), J=12, and M=5.}
    \label{fig:architecture}
\end{figure}

The overall pipeline from frontal view video of a walk to mean gait features is shown in Figure \ref{fig:overview}.
We take as input a monocular video of an individual walking toward the camera. We then apply a two-dimensional pose estimator to get a sequence of joint locations. We choose to use a two-dimensional pose estimator because we have found them to be more reliable than three-dimensional pose trackers on diverse clinical datasets, with less instances of catastrophic failure where the poses are meaningless. During training, we use the output of three different pose estimators \cite{cao_openpose_2021, fang_alphapose_2022, wu2019detectron2} as a form of data augmentation, but at test time we use AlphaPose \cite{fang_alphapose_2022}, as this model achieved the best performance in prior work \cite{sabo_concurrent_2022}. We then apply our gait analysis neural network, the architecture of which is shown in Figure \ref{fig:architecture}. 

This model predicts four gait features which have been shown to be clinically relevant in prior work: step length \cite{dolatabadi_quantitative_2018}, step width \cite{mehdizadeh_vision-based_2020, ng_measuring_2020, sabo_assessment_2020}, step time \cite{dolatabadi_quantitative_2018, ng_measuring_2020}, and velocity \cite{dolatabadi_quantitative_2018, sabo_assessment_2020}.
For each of these features, we predict the mean value over the walk, as the data used to train the model does not contain the information about the value of each feature for individual steps. Gait feature values are usually averaged before being fed into downstream clinical applications, so predicting the mean does not greatly limit clinical applicability of the features. 

\subsection{Dataset}
\label{sec:datasets}

The data used in this project was collected by the AMBIENT system as described by Dolatabadi et al. \cite{dolatabadi_feasibility_2019}. The system automatically detects participants walking down a hallway and records the walk with the video camera and the depth camera of a Kinect sensor. We use data collected from two sites to train and evaluate our model. Dataset 1 (DS1) was collected in a hospital setting, while Dataset 2 (DS2) was captured in a long term care facility. Data collection and analysis plans at each location were approved by the institutional ethics review boards. Gait features to be used as ground truth were extracted from the three dimensional depth data as done in a prior validation study that showed high agreement with values from the GAITRite\textsuperscript{\textregistered} electronic walkway \cite{DOLATABADI2016952}. Specifically, the derivative of the ankle keypoint trajectories was used to detect stance and swing times of each foot, based on the knowledge that a foot in contact with the ground is stationary. Once steps were detected, distance information for each step was then computed from the location of the ankle keypoints at each heelstrike event.
Table \ref{tab:datasets} summarizes the number of participants and walks captured in each location 
 and summary statistics about the selected gait features for each dataset. While most gait features show similar distributions across datasets, step length and velocity are higher for DS2, as participants in the long term care facility tended to be healthier and walk faster than those in the hospital setting.

\begin{table}
    \centering
    \caption{Summary of datasets used to train and evaluate the gait feature prediction neural network. Number of walks and participants are recorded, along with the mean value of each gait feature with the coefficient of variation in parentheses.
}
    \label{tab:datasets}
    \begin{tabular}{|c|c|c|c|c|c|c|}
    \hline
Dataset & Walks & Participants & Step Time & Step Width & Step Length & Velocity \tabularnewline
\hline
DS1 & 2216 & 37 & .60s (.21) & 17cm (.27) & 30cm (.32) & 50cm/s (.31) \tabularnewline
DS2 & 989 & 12 & .61s (.43) & 17cm (.27) & 32cm (.29) & 65cm/s (.33) \tabularnewline
Total & 3205 & 49 & .60s (.30) & 17cm (.27) & 31cm (.31) & 55cm/s (.34) \tabularnewline
\hline
\end{tabular} 

\end{table}

\section{Results} 

\subsection{Experimental Setup}
We performed the validation and evaluation of the model using 10-fold cross validation. As each dataset contained more than one walk from the same participant, we randomly assigned participants to one of the folds such that individuals from each dataset were evenly spread across folds. We then trained ten models, where each model was trained on eight folds, validated on one fold, and evaluated on one fold. Each model was trained for 200 epochs, and the model with the minimum weighted mean squared error on the validation set was used for evaluation. Results are aggregated over all the folds and then summary statistics are computed. 
We report Spearman's correlation coefficients ($\rho$) and mean average error (MAE) for each feature, and present results on each dataset individually as well as in aggregate to see if performance varies across the datasets. The Spearman's correlation coefficient shows if the predictions are consistent with the ground truth in terms of rank, while the mean average error shows how accurate the features are. A model with high correlation but lower mean average error likely has a bias, predicting higher or lower values than the ground truth.
We compare our model with the Transforming Gait model (TG) from Cotton et al. \cite{cotton_transforming_2022}, using the pre-trained model from that publication without any fine tuning. While retraining or fine tuning would be optimal, it requires per-step features which are much more difficult to obtain from our data than those averaged over the walk. Therefore, we use the model trained by the original authors on a diverse clinical dataset with a median age of 11 years and cerebral palsey and spina bifida as the most common conditions. The Transforming Gait model does not predict step width, so we compare on velocity, step length, and step time only. 

\subsection{Model Training}
We train our Pose2Gait model for 200 epochs. Each epoch uses all three versions of each walk predicted with each pose tracker as a form of data augmentation, randomizing the order over all sequences, with a batch size of 20 sequences. We use the Adam optimizer with learning rate $1^{-5}$ and apply the mean squared error loss with hand-crafted weights applied to each feature. The weights are necessary because the range of values for each feature varies, with step length being consistently larger than step width, for example. We chose weights to get the means of all features to be around 1, as shown in Table \ref{tab:feature_weights}. We normalize the input pose sequences by shifting and scaling to center the mid-hip point at (0, 0) and have a hip width of 1 at the center frame of the sequence. 
We choose 120 frames (4 seconds) as the input sequence length. We have seen that pose sequence quality is better when the participant closer to the camera, so rather than cropping randomly, we crop the full video to the portion at the end where the subject is closest to the camera as they are walking towards it. 

\begin{table}
    \small
    \centering
    \caption{Weights used when balancing loss across gait features. The weights were chosen to balance the contribution of the features to the overall loss. Without weighting, features with larger values, such as step length, would overpower features with smaller values, such as step width, when computing the gradient or determining the best model from the validation performance.} 
    \label{tab:feature_weights}
    \begin{tabular}{|c|c|c|c|c|}
    \hline
     Gait Feature & Step Time & Step Width & Step Length & Velocity   \\
     \hline
     Weight in Loss & 0.5 & 2.0 & 1.25 & 1.0  \\
     \hline
\end{tabular}
\end{table}

\subsection{Results} 
\begin{figure}
    \centering
    \includegraphics[width=\textwidth]{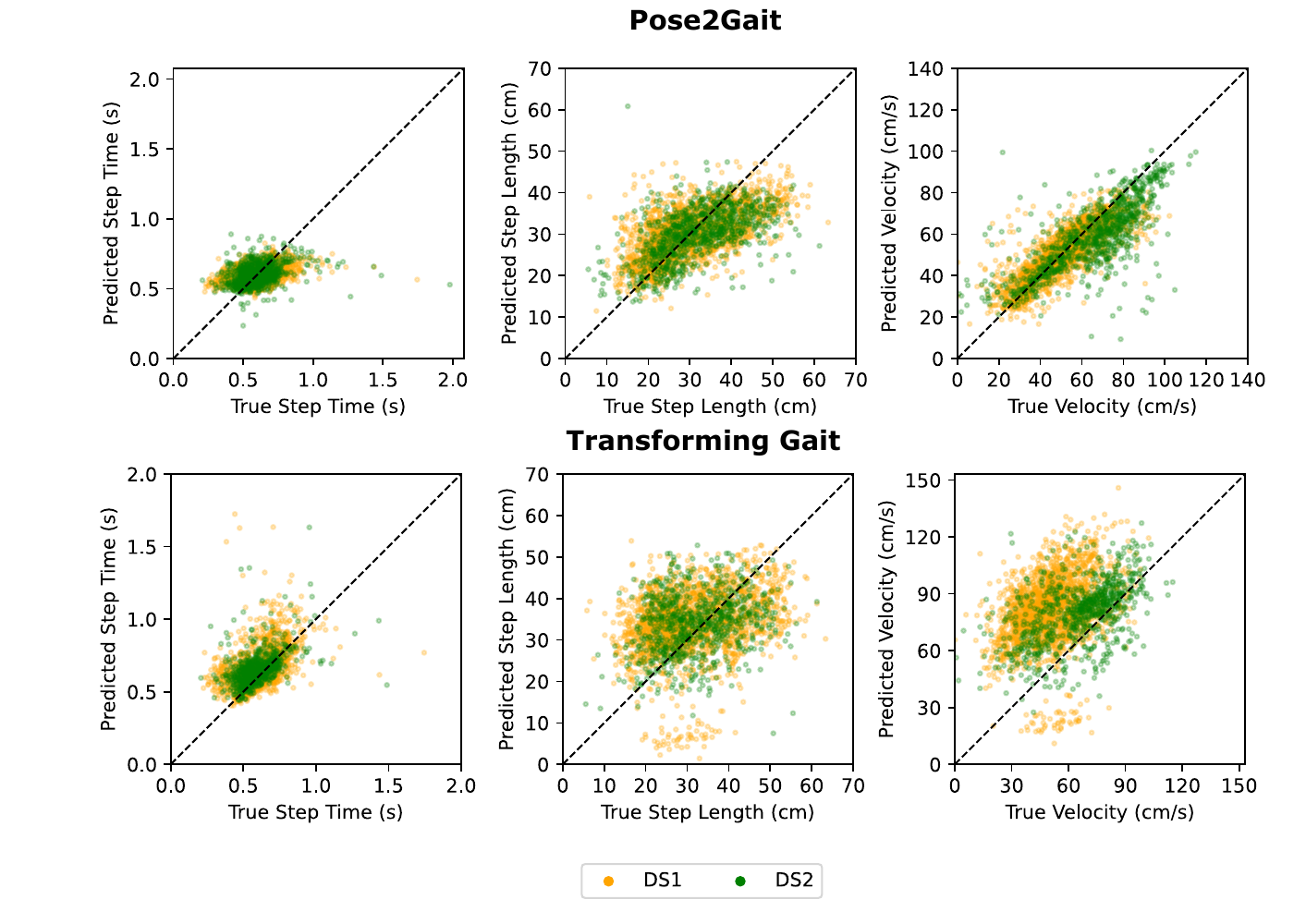}
    \caption{Correlation plots between true and predicted values for step time, step length, and velocity. Top row shows Pose2Gait predictions, while bottom row shows Transforming Gait predictions. Dataset membership is indicated by color of the point, and the equality line is plotted for reference.}
    \label{fig:correlation_plots}
\end{figure}

The overall results for our model Pose2Gait (P2G) and the Transforming Gait (TG) model \cite{cotton_transforming_2022} are reported in Table \ref{tab:overall_results}, with correlation plots for the common features shown in Figure \ref{fig:correlation_plots}. The Transforming Gait model failed to detect steps for 12.5\% of videos. The Transforming Gait model predicts the quadrature encoding of the gait phase and extracts the step times from zero crossing detection; therefore, if the predicted gait phase encodings did not cross zero (or a multiple of $2\pi$ after Kalman smoothing), no steps were detected. In the following tables and analysis, Transforming Gait results are presented only for the 87.5\% of videos where the model successfully obtained an output, potentially inflating results compared to Pose2Gait which obtained results for all videos. 

When examining the results aggregated across both datasets, the Pose2Gait model shows a high correlation with ground truth gait features for velocity (.81), a medium correlation for step length (.60), and low correlations for step time (.35) and step width (.27). Conversely, the Transforming Gait model achieves highest correlation for step time (.55), with low correlations on step length (.28) and velocity (.36). Similar trends are shown by the mean average error metric, with the Transforming Gait model showing lower error than Pose2Gait for step time and higher error for step length and velocity. Examination of the correlation plots in Figure \ref{fig:correlation_plots} shows different failure modes between the models. The Pose2Gait model predicts a relatively constant step time of around 0.6 seconds, failing to capture the variability of the data. The Transforming Gait model tends to overpredict the velocity, likely due to higher walking speeds in the population used to train the model.

The Pose2Gait results are surprising, as our input is two-dimensional pose sequences with individuals walking toward the camera, which we would expect to contain more information about the step width than the velocity and step length. These results show that the Pose2Gait neural network can extract three-dimensional gait information from a two-dimensional pose sequence, likely by examining the difference in size of the poses over time as the individual gets closer to the camera. However, our model struggles to extract step width, perhaps because we normalize the size of the pose sequence and do not provide the height of the participant, making it difficult to infer real world coordinates for features that are perpendicular to the camera.

\begin{table}
    \small
    \centering
    \caption{Spearman's correlation coefficient ($\rho$) and mean absolute error (MAE) for the Pose2Gait (P2G) and Transforming Gait (TG) models aggregated across all folds and datasets. The p-values associated with all Spearman's correlation coefficients in this table are less than 0.01, indicating statistical significance. Best result for each metric and feature is in bold.}
    \label{tab:overall_results}
    \begin{tabular}{|c|c|c|c|c|c|}
    \hline
        Metric & Model & Step Time & Step Width & Step Length & Velocity \\
        \hline
        \multirow{2}{*}{$\rho$ } & P2G & .35 & .27 & \textbf{.60}  & \textbf{.81}  \\
        & TG & \textbf{.55} & --- & .28 & .36 \\
        \hline
        \multirow{2}{*}{MAE} & P2G &  93ms & 3.4cm & \textbf{6.0cm} & \textbf{8.0cm/s} \\
        & TG & \textbf{88ms} & --- & 8.8cm & 28.1cm/s \\
        \hline
    \end{tabular}
\end{table}

In addition to examining the overall results, we show the correlation coefficients for DS1 only in Table \ref{tab:dataset1_results}, and DS2 in Table \ref{tab:dataset2_results}. As our datasets come from two locations with potentially different walking gaits, as shown by the difference in gait features between populations in Table \ref{tab:datasets}, we additionally include Pose2Gait models trained on each dataset individually (P2G-DS1 and P2G-DS2, respectively) to determine if training one global model or two separate models performs better. In each case, the Pose2Gait model trained on both datasets achieves the highest correlation for step length and velocity, while the Transforming Gait baseline performs best on step time, although the difference is slight for DS1. For both datasets, step width correlations improve slightly for the model trained on that dataset only. Mean average error results show similar trends to the correlation coefficients; these values are included in Appendix A for completeness.

\begin{table}
    \small
    \centering
    \caption{Spearman's correlation coefficients for DS1 only. The p-values associated with all coefficients in this table are less than 0.01, indicating statistical significance. Best result for each feature is in bold.}
    \label{tab:dataset1_results}
    \begin{tabular}{|c|c|c|c|c|}
    \hline
        Model & Step Time & Step Width & Step Length & Velocity \\
        \hline
        P2G & .35 & .30 & \textbf{.60}  & \textbf{.83}  \\
        P2G-DS1 & .32 & \textbf{.32} & .59 & .82 \\
        TG & \textbf{.36} & --- & .29 & .36 \\
        \hline
    \end{tabular}
\end{table}

\begin{table}
    \small
    \centering
    \caption{Spearman's correlation coefficients for DS2 only. The p-values associated with all coefficients in this table are less than 0.01, indicating statistical significance. Best result for each feature is in bold.}
    \label{tab:dataset2_results}
    \begin{tabular}{|c|c|c|c|c|}
    \hline
        Model & Step Time & Step Width & Step Length & Velocity \\
        \hline
        P2G & .35  & .18 & \textbf{.58}  & \textbf{.75}  \\
        P2G-DS2 & .22 & \textbf{.20} & .39 & .61 \\
        TG & \textbf{.53} & ---& .25 & .50 \\
        \hline
    \end{tabular}
\end{table}

\subsection{Ablation Study}
We change various aspects of our architecture and data augmentation and preprocessing to determine the effect on the accuracy of gait features predicted.
In addition to the Pose2Gait architecture described in this work, we tested the architecture from Lonini et al. \cite{lonini_2022} which incorporates batch normalization and max pooling layers in the convolutional architecture. As these features are standard for convolutional neural networks and the model accurately predicted temporal gait features from pose sequences in the original work, we were surprised to see that performance decreased significantly, with negative correlations for step time and step width, and only a .30 correlation for velocity. As such, we did not adopt this architecture and focused on variations of the Pose2Gait model described below.

First, many prior publications have used mirroring poses as a simple method of increasing the size of the training data. We trained the Pose2Gait+Mirror model with twice as many training samples by laterally reflecting each pose sequence. 
In addition, upper body joints tended to be noisier in our datasets due to some individuals walking with hands behind their backs. Since our gait features focus on the lower body, we trained a model with only the hips, knees, and ankles (Pose2Gait+Lower) to determine if removing the potentially extraneous upper body information improved performance. 
Previous work has shown that normalizing the pose sequences per frame, rather than per video as we have done in our main model, can be effective for predicting UPDRS scores \cite{sabo_estimating_2021}. While we suspected that normalizing per frame would remove crucial information about velocity and step length that is provided by the scale of the person as the move toward the camera, we trained the Pose2Gait+PerFrame model to test this normalization method.
Spearman's correlation coefficients for all these augmentations are shown in Table \ref{tab:spearman_results_ablation}, with the main Pose2Gait model shown for reference. Mean average error values are available in Appendix B. Although the mirror augmentation produced the best step time correlations and the lower body only model had the highest step width correlations, these positive effects came at the cost of other features. Per-frame normalization is the only augmentation that had an overall beneficial impact on all features, although the effect is slight. 
\begin{table}
    \small
    \centering
    \caption{Spearman's correlation coefficients for variations on the main Pose2Gait model.  The p-values associated with all coefficients in this table are less than 0.01, indicating statistical significance. Best result for each feature is in bold.}
    \label{tab:spearman_results_ablation}
    \begin{tabular}{|c|c|c|c|c|}
    \hline
        Model       & Step Time     & Step Width    & Step Length  & Velocity \\
        \hline
         P2G        &   .35         & .27           & \textbf{.60}  & .81  \\
         P2G+Mirror & \textbf{.41}  & .25           & \textbf{.60}  & .80 \\
         P2G+Lower &    .37         & \textbf{.44}  & .53           & .80 \\
         P2G+PerFrame &    .37      & .29           & \textbf{.60}  & \textbf{.83} \\
         \hline
    \end{tabular}
\end{table}

\section{Discussion} 
This paper presents a neural network that can predict velocity and step length from monocular video of individuals with dementia walking collected via environmental monitoring in a clinical setting. Results from the ablation study show that standard practices in computer vision for training convolutional networks, such as batch normalization, max pooling, and data augmentation, do not have a strictly positive effect when using convolutions to process joint locations instead of images. Similarly, the performance of the pre-trained Transforming Gait model suggests that gait analysis neural networks will not naturally generalize between different clinical populations, and fine-tuning or retraining is likely required. Additionally, architectures or data preprocessing techniques that work well for one gait analysis task might not be suitable for others, as shown by the architecture from Lonini et al. \cite{lonini_2022} working very well for temporal features in the original work but less well for spatiotemporal features here, as well as the inclusion of only lower body joints in the Pose2Gait model helping step width accuracy while hurting step length and velocity.


Future work remains to boost the accuracy of predictions for step time and step width. Because we have shown that different architecture and preprocessing approaches perform better for different features, training separate models for each feature could yield more accurate results, although it is much less efficient than sharing a single model.  To improve performance on a specific dataset, for example before deployment in an ambient monitoring setting, it would make sense to fine-tune the general Pose2Gait model on the target dataset only. Finally, while this work shows a proof of concept that three dimensional spatiotemporal gait features can be predicted from monocular video using computer vision techniques, gait features are only a proxy for the health of an individual. 
 Further work must be done to collect longitudinal data from ambient monitoring and to develop and validate clinical models that can detect sudden changes in gait that indicate high risk of negative health outcomes such as fall risk or hospitalization.

%
%
%
\bibliographystyle{splncs04}
\bibliography{references}
\newpage
\Large{\textbf{Appendix}}
\begin{appendix}
\section{Individual Dataset Results}
\label{app:dataset_maes}
\begin{table}[]
    \small
    \centering
    \caption{Mean average error values for DS1 only, including the main Pose2Gait model (P2G), the Pose2Gait model trained only on DS1 (P2G-DS1) and the baseline Transforming Gait model (TG). Best result for each feature is in bold.}
    \label{tab:dataset1_mae}
    \begin{tabular}{|c|c|c|c|c|}
    \hline
        Model & Step Time & Step Width & Step Length & Velocity \\
        \hline
        P2G &  89ms & \textbf{3.3cm} & \textbf{6.0cm} & \textbf{6.7cm/s} \\
        P2G-DS1 & 89ms & \textbf{3.3cm} & 6.2cm & 7.0cm/s \\
        TG & \textbf{87ms} & --- & 9.2cm & 32.0cm/s \\
        \hline
    \end{tabular}
\end{table}

\begin{table}[]
    \small
    \centering
    \caption{Mean average error values for DS2 only, including the main Pose2Gait model (P2G), the Pose2Gait model trained only on DS2 (P2G-DS2) and the baseline Transforming Gait model (TG). Best result for each feature is in bold.}
    \label{tab:dataset2_mae}
    \begin{tabular}{|c|c|c|c|c|}
    \hline
        Model & Step Time & Step Width & Step Length & Velocity \\
        \hline
        P2G &  102ms & \textbf{3.7cm} & \textbf{6.1cm} & \textbf{11.0cm/s} \\
        P2G-DS2 & 122ms & 3.9cm & 7.0cm & 14.8cm/s \\
        TG & \textbf{90ms} & --- & 7.8cm & 18.6cm/s \\
        \hline
    \end{tabular}
\end{table}

\section{Ablation Study Results}
\label{app:ablation_results}
\begin{table}[]
    \small
    \centering
    \caption{Mean average error values for all versions of the Pose2Gait model included in the ablation study: the main model (P2G), mirror augmentation (P2G+Mirror), lower body joints only (Pose2Gait+Lower), and per-frame normalization instead of per-video (P2G+PerFrame). Best result for each feature is in bold.}
    \label{tab:mae_results}
    \begin{tabular}{|c|c|c|c|c|}
    \hline
        Model       & Step Time     & Step Width    & Step Length       & Velocity \\
        \hline
        P2G         &  93ms         & 3.4cm         & \textbf{6.0cm}    & \textbf{8.0cm/s} \\
        P2G+Mirror  &  \textbf{89ms} & 3.5cm         & 6.1cm             & 8.3cm/s \\
        P2G+Lower   &  94ms         & \textbf{3.2cm}   & 6.4cm           & 8.5cm/s \\
        P2G+PerFrame & 91ms        &   3.4cm         & 6.1cm             & \textbf{8.0cm/s} \\
        \hline
    \end{tabular}
\end{table}
\end{appendix}

\end{document}